\documentclass[conference]{IEEEtran}
\usepackage{amsmath,amsfonts,amssymb}
\usepackage{graphicx}
\usepackage[caption=false,font=footnotesize,labelfont=sf,textfont=sf]{subfig}
\usepackage{booktabs,tabularx,makecell,array,multirow}
\usepackage{cite,url,stfloats,verbatim,xcolor}
\newcolumntype{Y}{>{\centering\arraybackslash}X}

\def\BibTeX{{\rm B\kern-.05em{\sc i\kern-.025em b}\kern-.08em
    T\kern-.1667em\lower.7ex\hbox{E}\kern-.125emX}}
\usepackage{algorithm}
\usepackage{algpseudocode}

\begin{document}
\title{Trust-free Personalized Decentralized Learning\\
\thanks{\textsuperscript{*}Corresponding author.}
}

\author{
    \IEEEauthorblockN{
        Yawen Li, 
        Yan Li, 
        Junping Du, 
        Yingxia Shao, 
        Meiyu Liang and 
        Guanhua Ye\textsuperscript{*}
    }
    \IEEEauthorblockA{\textit{Beijing University of Posts and Telecommunications}, Beijing, China}
    \IEEEauthorblockA{\{warmly0716, liyanly, junpingd, shaoyx, meiyu1210, g.ye\}@bupt.edu.cn}
}

\maketitle

\begin{abstract}
Personalized collaborative learning in federated settings faces a critical trade-off between customization and participant trust. Existing approaches typically rely on centralized coordinators or trusted peer groups, limiting their applicability in open, trust-averse environments. While recent decentralized methods explore anonymous knowledge sharing, they often lack global scalability and robust mechanisms against malicious peers. To bridge this gap, we propose TPFed, a \textit{Trust-free Personalized Decentralized Federated Learning} framework. TPFed replaces central aggregators with a blockchain-based bulletin board, enabling participants to dynamically select global communication partners based on Locality-Sensitive Hashing (LSH) and peer ranking. Crucially, we introduce an ``all-in-one'' knowledge distillation protocol that simultaneously handles knowledge transfer, model quality evaluation, and similarity verification via a public reference dataset. This design ensures secure, globally personalized collaboration without exposing local models or data. Extensive experiments demonstrate that TPFed significantly outperforms traditional federated baselines in both learning accuracy and system robustness against adversarial attacks.
\end{abstract}

\begin{IEEEkeywords}
Federated Learning, 
Personalized Decentralized Learning, 
Trust-free Collaboration, 
Knowledge Distillation.
\end{IEEEkeywords}

\section{Introduction}
The proliferation of distributed devices has established Federated Learning (FL)~\cite{bonawitz2019towards} as a standard for privacy-preserving intelligence. In broader networked intelligent systems, robust operation under uncertainty and time-varying delays has long been recognized as a prerequisite for reliable coordination~\cite{li2008robust}. While conventional Centralized FL (CFed) aggregates updates globally, it often struggles with heterogeneous, non-IID data distributions common in real-world scenarios~\cite{wang2020optimizing}. Consequently, personalized FL has emerged, where clients adapt a global model to local distinctiveness. Recent studies emphasize that constructing a \emph{communication graph}—allowing clients to selectively exchange information only with high-quality, similar neighbors—yields superior personalized outcomes~\cite{pmlr-v202-zhang23a}. Graph-based FL studies, such as cross-graph node classification, further suggest that collaboration should account for structural relations among clients~\cite{guan2021federated}.

However, implementing such selective collaboration without a trusted third party remains a significant challenge. Existing Semi-decentralized (SFed) methods~\cite{sun2023semi} employ a central coordinator to manage topology, introducing a single point of failure and inherent trust assumptions. Conversely, fully Decentralized (DFed) approaches~\cite{beltran2023decentralized} typically rely on local gossip protocols. These often suffer from limited global visibility and lack rigorous mechanisms to verify peer honesty, leaving the system vulnerable to poisoning or free-riding behaviors.

To address these limitations, we propose \textbf{TPFed} (\textit{Trust-free Personalized Decentralized Federated Learning}). TPFed adheres to a strict trust-free design philosophy: no single entity possesses control, and no participant is assumed to be inherently trustworthy. Unlike conventional gossip-based schemes (DFed(G)), TPFed introduces a ``Web-based'' paradigm (DFed(W)) that enables global neighbor discovery and verification through three integrated mechanisms:

\begin{figure*}[t] 
\centering
\includegraphics[scale=0.25]{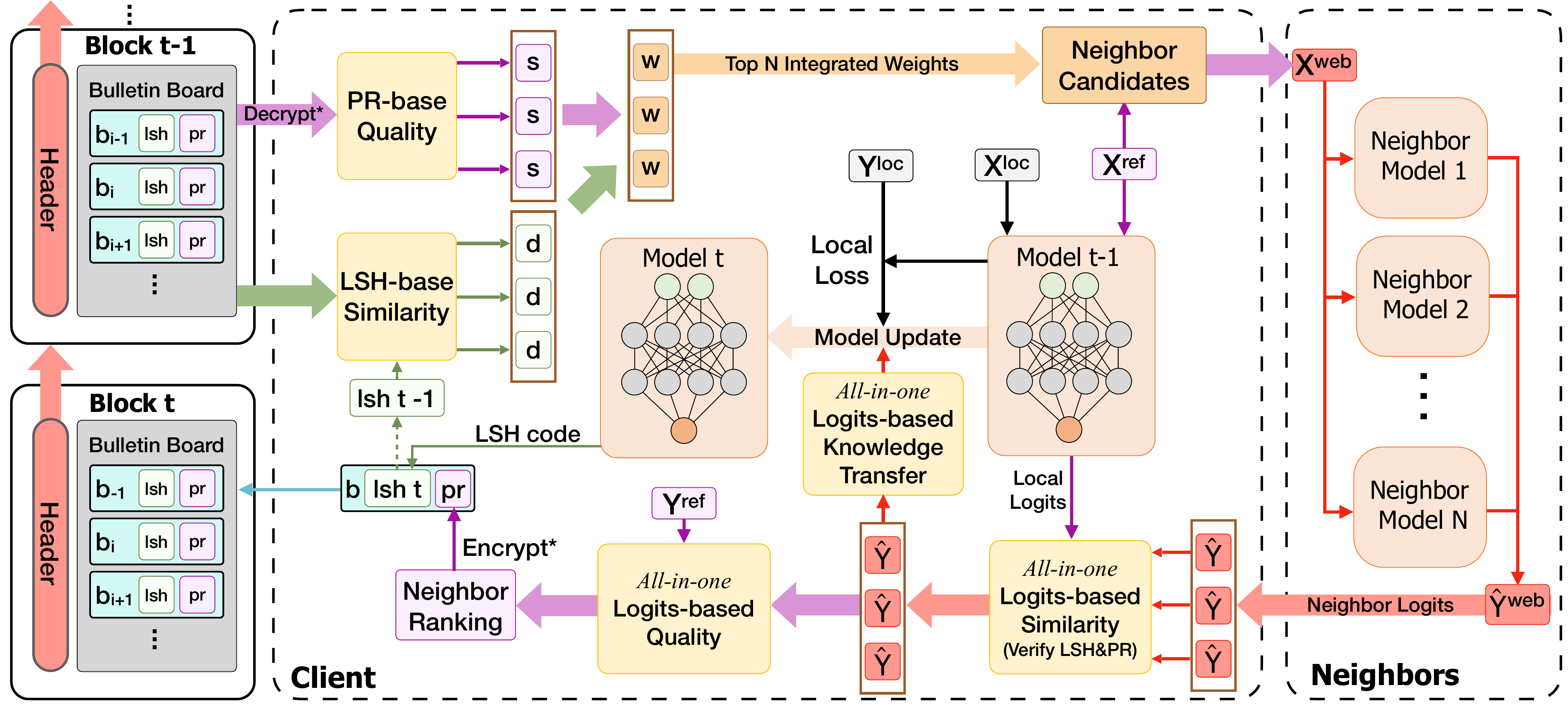}
\caption{\textbf{Overview of the TPFed framework.} The central coordinator is replaced by a blockchain-based bulletin board, which filters neighbor candidates based on inter-client similarity (\textit{via LSH}) and model quality (\textit{via PR}). Clients then employ \emph{all-in-one logits} to simultaneously perform knowledge transfer, model quality evaluation, and inter-client similarity measurement, where the latter also verifies the LSH and PR claims. ${}^{*}$In practice, clients publish commitments $C$ in the bulletin instead of the original neighbor ranking \textit{PR};  however, \textit{PR} is depicted in the figure to simplify the illustration of data flow.}  
\label{fig_overview}
\vspace{-15pt}
\end{figure*}

\textbf{All-in-one P2P Knowledge Distillation.} TPFed extends knowledge distillation to a fully decentralized setting, allowing each client to perform knowledge transfer, model quality evaluation, and inter-client similarity measurement \emph{in one shot}. Specifically, each client requests from its peers the predicted logits for a locally held (and privacy-preserving) reference dataset~\cite{zhou2024federated,wan2024research}. This eliminates the need for trusting a central aggregator with raw model updates or sensitive data, ensuring the process is inherently trust-free.

\textbf{Immutable Bulletin Board for Coordination.} 
Instead of relying on a centralized coordinator, TPFed leverages a public immutable ledger (e.g., a smart contract or a Layer-2 blockchain solution) to serve as a global bulletin board. This guarantees that no third party can manipulate or censor posted bulletins. To ensure practical scalability and low latency, we adopt an asynchronous update scheme: clients retrieve neighbor information from the ledger but only commit their own updates periodically (e.g., every $K$ rounds) or when significant model changes occur. By simply reading the transparent ledger, each client discovers with whom to communicate via Locality-Sensitive Hashing (LSH)~\cite{kollias2022sketch,chen2022feddual} and Peer Ranking (PR) mechanisms. This design removes the single point of failure while maintaining a trust-free coordination layer.

\textbf{Countermeasures Against Malicious Behaviors.} The trust-free nature of open networks invites adversaries to forge LSH codes or collude. TPFed secures the system via a commit-then-reveal ranking scheme and a fine-grained consistency check. By comparing a peer's claimed LSH distance against the actual divergence of their logits, clients can locally detect and blacklist malicious actors who attempt to spoof similarity or degrade performance.

In light of this, we differentiate TPFed from other fully decentralized federated learning approaches. In the sections that follow, we refer to conventional decentralized FL reliant on \textbf{G}ossip-based communication as DFed(\textbf{G}), whereas our TPFed framework is denoted as DFed(\textbf{W}), reflecting its ability to retrieve blockchain information via the \textbf{W}eb and perform direct peer-to-peer exchanges with any node. We believe this Web3.0-oriented FL paradigm holds significant potential for future research. Our main contributions are summarized as follows:

\begin{itemize}
\item We identify the gap in achieving robust \emph{model quality} and \emph{similarity evaluation} in fully decentralized, trust-free FL environments.
\item We propose TPFed, a novel framework combining blockchain-based coordination with all-in-one knowledge distillation to enable secure, global-scale neighbor selection.
\item Comprehensive experiments on real-world datasets demonstrate that TPFed substantially outperforms existing baselines in accuracy and resilience against DoS and poisoning attacks.
\end{itemize}

\section{Related Work}
\textbf{Decentralized Federated Learning.}
Decentralized Federated Learning (DFL) mitigates the single-point-of-failure risks associated with centralized servers by enabling direct client-to-client coordination~\cite{abdellatif2024sdcl}. This architecture is particularly advantageous for privacy-sensitive and communication-constrained environments. Communication-efficient FL, including stochastic quantization, has been explored to reduce transmission cost, but it does not resolve decentralized neighbor verification~\cite{li2022federated}. However, the majority of DFL approaches rely on static or locally restricted topologies, such as physical proximity or random gossip protocols. While these methods function in stable settings, they often underutilize global network information. Consequently, they lack the adaptability required for dynamic environments where optimal collaboration partners may not be topologically adjacent~\cite{hijazi2024collaborative}.

\textbf{Personalized Neighbor Selection Mechanism.}
Effective neighbor selection is pivotal for achieving personalization and fast convergence in DFL~\cite{10.1145/3543507.3583212}. Conventional solutions typically employ heuristic strategies, such as location-based clustering~\cite{domini2024proximity}, graph community detection~\cite{yang2016modularity}, or random sampling~\cite{ye2022decentralized}, which fail to account for the semantic similarity between models. To address this, recent works have leveraged Locality-Sensitive Hashing (LSH) to estimate model similarity while preserving privacy~\cite{chen2022feddual}. Nevertheless, relying exclusively on LSH without verifying model quality or preventing adversarial manipulation remains insufficient for ensuring robust personalization in open networks. Related recommendation studies have used self-supervised graph co-training and filter-enhanced MLP architectures to improve personalized sequence modeling~\cite{xia2021self,zhou2022filter}; however, these methods do not provide a trust-free global neighbor-selection protocol for DFL.

\textbf{Web-based Collaborative Learning.}
The evolution of web technologies, including WebRTC~\cite{blum2021webrtc}, Distributed Hash Tables (DHTs)~\cite{rossi2022webdht}, and Decentralized Identifiers (DIDs)~\cite{dib2020decentralized}, has paved the way for serverless, real-time collaboration. These open architectures, however, introduce significant trust and security challenges, prompting a shift toward blockchain-facilitated frameworks to enforce accountability~\cite{deng2020trust}. Furthermore, to safeguard data during these interactions, cryptographic techniques such as differential privacy~\cite{zhao2022survey} and secure multi-party computation~\cite{olakanmi2022trust} are increasingly integrated into collaborative pipelines.

\section{Preliminaries}
This section introduces the key concepts and components of TPFed: user clients, bulletins, peer-to-peer (P2P) communication, and the system-wide objective.

\subsection{User Clients}\label{sec_P_client}
Each user $i$ corresponds to a client $u_i$ that maintains a local dataset $(\mathbf{X}^{\text{loc}}_i,\mathbf{Y}^{\text{loc}}_i)$, a reference dataset $(\mathbf{X}^{\text{ref}}_i,\mathbf{Y}^{\text{ref}}_i)$, and local parameters $\theta_i$. The inference of model $\theta_j$ is $f(\theta_j,\cdot)$. The reference dataset facilitates communication with others without exposing sensitive local data. Each client manages a dynamic set of neighbors $\mathcal{N}_i$, which may change over time. Communication with these neighbors occurs in a P2P manner.

\subsection{Bulletins}\label{sec_P_bulletins}
Each client publicizes a \emph{bulletin} $\mathbf{b}_i^{(t)} = \{\text{lsh}_i^{(t)}, \mathbf{R}_i^{(t)}\}$ on the blockchain, where $\text{lsh}_i^{(t)}$ is a locality-sensitive hash (LSH) of the model at iteration $t$, and $\mathbf{R}_i^{(t)}$ ranks its neighbors based on performance evaluations with the reference dataset
\footnote{Practically implemented as a commitment $C_i^{(t)}$ (see Sec.~\ref{sec_M_countermeasures}); $\mathbf{R}_i^{(t)}$ is used here to simplify the scoring formulation.}. 
From the bulletins $\{\mathbf{b}_j^{(t)} | j\neq i\}$, client $u_i$ infers similarity scores $\{d_{ij}\}$ from the LSH codes and ranking scores $\{s_j\}$ from peer performance. This eliminates the need for a central coordinator and allows autonomous neighbor selection.

\subsection{Dynamic Communication Graph}\label{sec_P_graph}
The overall network is a dynamic graph $G=(U,E,W)$ with $U=\{u_i|i\in[1,M]\}$, where $M$ is the total number of clients. An edge $e_{ij}=1$ indicates direct P2P connectivity between $u_i$ and $u_j$, and $w_{ij}$ represents the potential benefit of their communication. Typically, all clients can communicate ($e_{ij}=1$), while $w_{ij}$ is derived from bulletins and reflects model similarity and peer performance.

\subsection{TPFed Objective}\label{sec_P_objective}
TPFed aims to help each client find an optimal neighbor set $\mathcal{N}_i$ by leveraging $W$ to improve overall performance. The global objective is:
\begin{equation}\label{eq:global_obj}
\resizebox{0.9\linewidth}{!}{$ 
\begin{aligned}
    \mathcal{L}_{\text{global}}^* &= \frac{1}{M}\sum_{i=1}^M \Bigl[ \alpha\,\ell\bigl(f(\theta_i,\mathbf{X}^{\text{loc}}_i),\mathbf{Y}^{\text{loc}}_i\bigr) + (1-\alpha)\,\mathcal{L}_i^{\text{ref}}
    \Bigr],
\end{aligned}
$}
\end{equation}
where $\mathcal{L}_i^{\text{ref}}$ measures alignment with neighbors on $\mathbf{X}^{\text{ref}}_i$, $\ell(\cdot)$ is the loss function, and $\alpha\in[0,1]$ controls the balance between local and collaborative terms. By minimizing $\mathcal{L}_{\text{global}}^*$, TPFed encourages robust local performance while promoting effective cooperation across the network.

\section{Methodology}
This section details the TPFed framework. The complete workflow is summarized in Algorithm~\ref{alg:tpfed_client}.

\subsection{All-in-one P2P Knowledge Distillation} \label{sec_M_distillation}

\begin{figure}[h] 
\centering
\vspace{-10pt}
\includegraphics[scale=0.235]{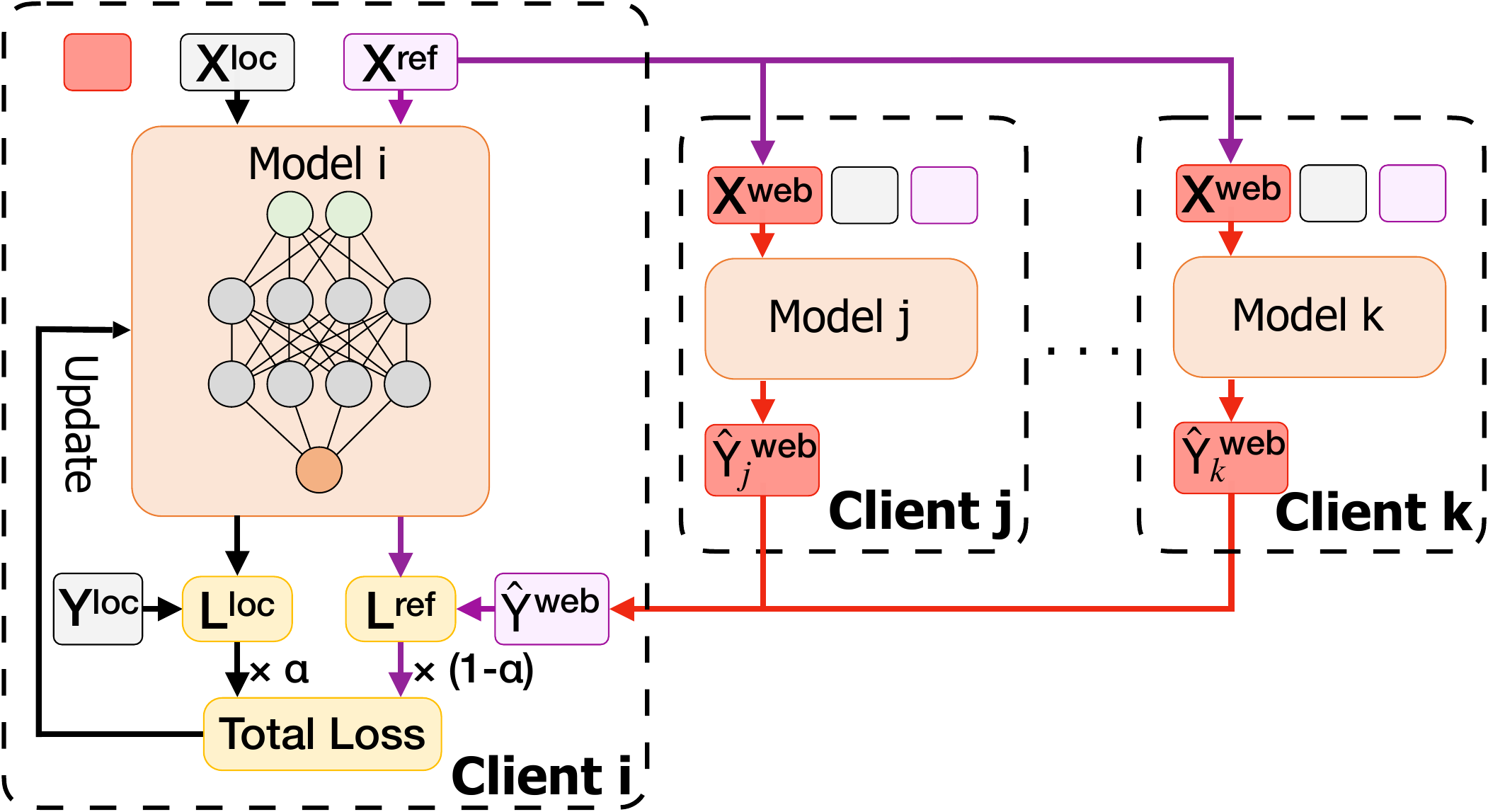}
\vspace{-5pt}
\caption{Client learns from neighbors via collecting logits $\hat{Y}^{\text{web}}$.}  
\label{fig_p2p_comm}
\vspace{-5pt}
\end{figure}

\textbf{Knowledge Transfer.} Clients in TPFed communicate directly to refine local models via knowledge distillation. 
To strictly preserve privacy, the data exchanged during this process comes from a \emph{public, non-sensitive reference dataset} $\mathcal{D}^{\text{ref}}$, which is disjoint from any client's private local data. 
As shown in Fig.~\ref{fig_p2p_comm}, the process starts with client $u_i$ sharing the features $\mathbf{X}^{\text{ref}}_i$ (sampled from $\mathcal{D}^{\text{ref}}$) with neighbor $u_j$, which responds with $\hat{Y}^{\text{web}}_j = f(\theta_j, \mathbf{X}^{\text{ref}}_i)$. 
Although $\mathbf{X}^{\text{ref}}_i$ follows a public distribution different from the private non-IID data, it serves as a semantic anchor~\cite{duan2023federated}. This use of a shared anchor is consistent with classical multi-source fusion studies, where complementary signals are first mapped into a common representation before integration~\cite{xu2013image}. If two models produce similar logits on a diverse public dataset, they likely share compatible decision boundaries, making knowledge transfer effective even across heterogeneous domains~\cite{makhija2022architecture}.
Client $u_i$, having the ground truth $\mathbf{Y}^{\text{ref}}_i$ (from the public source), assesses $u_j$’s performance and updates $\theta_i$ by aligning $f(\theta_i, \mathbf{X}^{\text{ref}}_i)$ with these logits. When multiple neighbors participate, the update aggregates all $\hat{Y}^{\text{web}}_j$:

\begin{equation}\label{eq:transfer}
\mathcal{L}_i^{\text{ref}} 
=  \Bigl\| f\bigl(\theta_i,\mathbf{X}^{\text{ref}}_i\bigr) 
  - \tfrac{1}{N}\textstyle\sum_{j=1}^N  \hat{Y}^{\text{web}}_j \Bigr\|^2.
\end{equation}

\textbf{All-in-one nature.} In a single exchange of logits, TPFed simultaneously achieves knowledge transfer, peer quality measurement, and similarity verification. By comparing each neighbor's $\hat{Y}^{\text{web}}_j$ against $\mathbf{Y}^{\text{ref}}_i$, the client quantifies that neighbor’s model quality, while examining how well different logits align with the local predictions provides a measure of inter-model similarity. This approach keeps reference data private and cuts communication overhead, as the same logits serve both local updates and evaluation needs.

\subsection{Personalization via Bulletin Board} \label{sec_M_personal}

Semi-decentralized methods benefit from a globally informed coordinator that can select neighbors based on inter-client similarity and model quality. In TPFed, we replace this centralized coordinator with a blockchain-based bulletin board to achieve a similar effect.

\textbf{LSH-based similarity.} Each client posts a \emph{locality-sensitive hash} (LSH) code derived from its model parameters. To mitigate the high cost and latency associated with frequent blockchain interactions, clients are not required to update their LSH codes in every iteration. Instead, we implement an \emph{asynchronous update strategy} with an interval $T_{\text{sync}}$. Client $u_i$ computes and publishes $\text{lsh}_i = \text{LSH}(\theta_i)$ every $T_{\text{sync}}$ rounds. Between updates, peers rely on the most recent on-chain snapshot for neighbor selection. Models that are highly similar generate hash codes that match with high probability. The Hamming distance between two codes,The Hamming distance between two codes,
\begin{equation}\label{eq_dist}
    d_{ij} = \text{HammingDist}\bigl(\text{lsh}_i, \text{lsh}_j\bigr),
\end{equation}
reflects their approximate similarity; a smaller $d_{ij}$ indicates closer models. 
It is worth noting that while neural networks are permutation invariant, 
we adopt the standard FL assumption of a shared initialization $\theta^{(0)}$. 
This ensures that client models remain in a broadly aligned parameter space during training, making LSH on parameters a computationally efficient proxy for functional similarity. 
Even in cases where permutation leads to falsely large distances, TPFed's redundancy allows clients to find other sufficient neighbors, 
while the subsequent logits verification prevents false positives\cite{li2013gaussian}.

\textbf{PR-based quality.} Each client also publishes a Peer Ranking (PR) based on the peer losses observed during P2P communication. For example, $u_i$ produces an ordered list $\mathbf{R}_i$, where higher-ranked neighbors have lower loss $\ell_{ij}$. By aggregating these rankings from all clients, each peer $u_j$ is assigned a \emph{ranking score}:
\begin{equation}\label{eq_rank}
    \small 
    s_j = \frac{\bigl|\{ \mathbf{R}_k \mid u_j \in \text{top-}K(\mathbf{R}_k),\, k \in [1,M]\}\bigr|}
               {\bigl|\{\mathbf{R}_k \mid u_j \in \mathbf{R}_k,\, k \in [1,M]\}\bigr|}
\end{equation}
This score captures how often $u_j$ is a top-$K$ performer across the network.

\textbf{Personalized neighbor selection.} Each client combines the ranking score $s_j$ and similarity distance $d_{ij}$ to derive a weight:
\begin{equation}\label{eq_weight}
    w_{ij} = s_j \cdot \exp\bigl(-\gamma \cdot d_{ij}\bigr),
\end{equation}
where $\gamma$ is a hyperparameter controlling the emphasis on similarity. A higher weight indicates a neighbor that is both well-ranked and closely aligned in model space. Client $u_i$ then selects the top $N$ peers with the largest $w_{ij}$ values to form its neighbor set $\mathcal{N}_i$, ensuring collaborations are tailored to each client’s unique observations and requirements.

\textbf{Exploration Mechanism.} Relying strictly on $w_{ij}$ may prevent unranked peers from ever being evaluated (the ``cold start'' problem). To address this, we employ an $\epsilon$-greedy strategy: client $u_i$ selects $(1-\epsilon)N$ neighbors based on the highest weights (exploitation) while randomly sampling the remaining $\epsilon N$ candidates from the network.

\begin{algorithm}[t]
\caption{TPFed Framework for Client $u_i$}\label{alg:tpfed_client}
\footnotesize
\begin{algorithmic}[1]
\Require Local $\{\mathbf{X}_i^{\text{loc}},\mathbf{Y}_i^{\text{loc}}\}$, Reference $\{\mathbf{X}_i^{\text{ref}},\mathbf{Y}_i^{\text{ref}}\}$, Init $\theta_i$, Iterations $T$, Hyperparams $(\alpha_i,\gamma, \epsilon)$, Neighbors $N$
\For{$t=1$ \textbf{to} $T$}
  \State \textbf{Neighbor selection}
  \If{$t>1$}
    \State Retrieve $\mathbf{b}_j^{(t-1)}=\{\text{lsh}_j^{(t-1)},C_j^{(t-1)}\}$ and $\mathbf{R}_j^{(t-1)}$; verify $\mathbf{R}_j^{(t-1)}$
  \EndIf
  \State For each $u_j$: compute $d_{ij}$ (Eq.~(2)) and $s_j$ (Eq.~(3))
  \State $w_{ij}\gets s_j\exp(-\gamma d_{ij})$; $\mathcal{N}_i^{(t)}\gets$ Top-$(1\!-\!\epsilon)N$ by $w_{ij} \cup$ Random-$\epsilon N$

  \State \textbf{Communication}
  \For{$u_j\in\mathcal{N}_i^{(t)}$}
    \State Send $\mathbf{X}_i^{\text{ref}}$ to $u_j$, receive $f(\theta_j^{(t)},\mathbf{X}_i^{\text{ref}})$
    \State $\ell_{ij}^{(t)}\gets \ell\!\left(f(\theta_j^{(t)},\mathbf{X}_i^{\text{ref}}),\mathbf{Y}_i^{\text{ref}}\right)$
  \EndFor
  \State \textbf{Model update}
  \State Calculate weights $p_{ij} \propto w_{ij}^{\text{final}}$ (Eq.~6) using $\ell_{ij}^{(t)}$ and consistency $\mathcal{V}_{ij}$
  \State $\bar f^{(t)}\!\gets\! \sum_{u_j\in\mathcal{N}_i^{(t)}} p_{ij} \cdot f(\theta_j^{(t)},\mathbf{X}_i^{\text{ref}})$
  \State $\mathcal{L}^{(t)}\!\gets\! \alpha_i\,\ell\!\big(f(\theta,\mathbf{X}_i^{\text{loc}}),\mathbf{Y}_i^{\text{loc}}\big)+(1-\alpha_i)\!\left\|f(\theta,\mathbf{X}_i^{\text{ref}})-\bar f^{(t)}\right\|^2$
  \State $\theta_i^{(t+1)}\gets\text{Update}\big(\theta_i^{(t)},\nabla_\theta \mathcal{L}^{(t)}\big)$
  \State \textbf{Bulletin publication}
  \State $\text{lsh}_i^{(t)}\gets$ update via Eq.~(4); $\mathbf{R}_i^{(t)}\gets$ sort by ascending $\ell_{ij}^{(t)}$
  \State $C_i^{(t)}\gets \text{Hash}(\mathbf{R}_i^{(t)})$; publish $\mathbf{b}_i^{(t)}=\{\text{lsh}_i^{(t)},C_i^{(t)}\}$
  \State Reveal $\mathbf{R}_i^{(t)}$ for verification at $t\!+\!1$
\EndFor
\end{algorithmic}
\end{algorithm}

\subsection{Countermeasures Against Malicious Behaviors} \label{sec_M_countermeasures}The trust-free nature of TPFed introduces two primary attack vectors: \emph{Sybil/DoS attacks} via LSH forgery, and \emph{Collusion/Ranking manipulation} in peer reviews. We address these through a rigorous verify-and-penalize mechanism that leverages the intrinsic workflows of TPFed.

\textbf{LSH Forgery \& DoS.} Adversaries may forge LSH codes to minimize the Hamming distance $d_{ij}$ (Eq.~\ref{eq_dist}), tricking client $u_i$ into selecting them as neighbors. A naive system would be vulnerable to Denial-of-Service (DoS) attacks, where checking massive numbers of fake candidates exhausts $u_i$'s bandwidth. We repurpose this existing data flow to perform a \emph{strict consistency check}. Upon receiving logits $\hat{Y}_j^{\text{web}}$ from a selected neighbor $u_j$, client $u_i$ computes the actual semantic divergence $D_{\text{KL}}$ and compares it with the claimed LSH distance:

\begin{equation}\label{eq:consistency}
\mathcal{V}_{ij} = \mathbb{I}\left( \left| \frac{d_{ij}}{L} - \phi(D_{\text{KL}}(\mathbf{Y}_i^{\text{ref}} || \hat{Y}_j^{\text{web}})) \right| \le \tau \right),
\end{equation}
where $L$ is the hash length, $\phi(\cdot)$ is a normalizing mapping function, and $\tau$ is a tolerance threshold.If $\mathcal{V}_{ij} = 0$ (check failed), it implies $u_j$'s LSH code contradicts its actual model behavior. To prevent DoS recurrence, $u_i$ immediately adds $u_j$ to a \textbf{Local Blacklist} for $T_{\text{ban}}$ rounds.This creates an \emph{asymmetric cost structure}: attackers must pay real-world blockchain transaction fees to continuously register new identities (LSHs) to sustain an attack, whereas honest clients filter them out with a one-time verification cost.

\textbf{Collusion.} Malicious cliques may engage in \emph{collusion}, mutually boosting each other's ranking scores $s_j$.\emph{Countermeasure:} In Sec.~\ref{sec_M_personal}, we introduced a \textbf{commit-and-reveal} scheme (publishing hash commitments $C_i^{(t)}$) to prevent clients from altering their evaluations after seeing others'. While this prevents ex-post modification, it cannot verify the \emph{honesty} of the ranking source itself (e.g., pre-planned collusion).TPFed resolves this by treating the global ranking score $s_j$ merely as a \emph{discovery heuristic}, not a ground truth. The final decision weight $w_{ij}$ is dominated by the \emph{local} evaluation of the exchanged logits. Even if a malicious node $u_j$ achieves a high $s_j$ through collusion, its effective weight is dynamically adjusted:
\begin{equation}w_{ij}^{\text{final}} = \mathcal{V}_{ij} \cdot w_{ij} \cdot \exp\left( - \lambda \cdot \mathcal{L}_i^{\text{ref}}(u_j) \right).
\end{equation}
If the locally measured loss $\mathcal{L}_i^{\text{ref}}(u_j)$ is high (indicating poor alignment with $u_i$'s objective), the neighbor is effectively discarded regardless of its inflated global reputation. This ensures that collusion attacks can at most affect the \emph{candidate selection} phase but cannot corrupt the actual \emph{model update}.

\textbf{Reference Overfitting.}
Besides, to prevent advanced adversaries from overfitting to the static public reference dataset $\mathbf{X}^{\text{ref}}$ (i.e., memorizing answers to pass checks while holding a corrupted model), clients employ \emph{Randomized Querying}. In each round, $u_i$ applies random augmentations (e.g., cropping, noise injection) to $\mathbf{X}^{\text{ref}}$ before sending it to neighbors. This forces the adversary to provide logits that are robust to unseen perturbations. Crucially, if an adversary possesses a model robust enough to handle these random augmentations consistently, their contribution becomes mathematically equivalent to that of a benign, high-quality teacher. In this scenario, the attack becomes self-defeating: to pass the verification, the adversary is forced to behave honestly and contribute valid knowledge, thereby nullifying the intent to degrade the victim's performance.

\section{Experiments}

In this section, we conduct comprehensive experiments to evaluate the TPFed framework. We structure our evaluation to answer the following Research Questions (RQs):

\begin{itemize}
    \item \textbf{RQ1 (Effectiveness):} Does TPFed outperform existing centralized, semi-decentralized, and decentralized federated learning baselines in terms of model accuracy?
    \item \textbf{RQ2 (Identity Defense):} Can TPFed effectively detect and reject adversaries attempting to spoof model similarity (LSH forgery) to perform impersonation attacks?
    \item \textbf{RQ3 (Poisoning Resilience):} How robust is TPFed against active poisoning attacks where malicious clients inject noise to disrupt the training process?
    \item \textbf{RQ4 (Scalability):} How does the communication overhead of TPFed compare to other methods, and does the TPFed-idx variant effectively improve scalability?
    \item \textbf{RQ5 (Ablation):} What is the contribution of each core component (LSH, Peer Ranking, and Logits Verification) to the overall framework performance?
    \item \textbf{RQ6 (System Robustness):} Can TPFed maintain stability under adverse system conditions, such as network delays, transmission failures, or low-quality reference data?
\end{itemize}

\subsection{Baselines}
We compare \textbf{TPFed} with four representative baselines covering the spectrum from isolated training to fully decentralized learning.

\noindent\textbf{SILO}~\cite{lian2017can}.  
Devices train solely on local data without any interaction. This maximizes privacy but loses the benefits of cross-device collaboration, typically yielding weaker models.

\noindent\textbf{Centralized (CFed).}  
We include \textbf{BlockFL}~\cite{lu2019blockchain} and \textbf{FedMD}~\cite{Li2019FedMD}. BlockFL employs a blockchain-backed mechanism to share gradients and build a global model but still depends on a central coordinator. FedMD aggregates heterogeneous models via distillation using a public or proxy dataset under centralized orchestration.

\noindent\textbf{Semi-Decentralized (SFed).}  
We evaluate \textbf{pFedClub}~\cite{wang2024pfedclub} and \textbf{SQMD}~\cite{ye2023heterogeneous}. pFedClub personalizes models through functional block decomposition, while SQMD manages asynchronous participation and device heterogeneity using a central communication graph. Both achieve strong performance but require trusted coordinators, limiting robustness in fully decentralized environments.

\noindent\textbf{Decentralized (DFed(G)).}  
We consider \textbf{ProxyFL}~\cite{kalra2023decentralized} and \textbf{KD-PDFL}~\cite{jeong2023personalized}. ProxyFL exchanges proxy models for enhanced privacy, and KD-PDFL uses knowledge distillation for peer collaboration. Although free of central aggregators, these methods still assume honest peer behavior and remain vulnerable to adversarial manipulation.

\begin{table*}[t]
\centering
\caption{Performance Comparison of Different Frameworks Across Base Models and Datasets}
\label{tab:performance}
\small
\renewcommand{\arraystretch}{0.85}
\setlength{\tabcolsep}{10pt} 
\begin{tabular}{llcccc}
\toprule
\textbf{Category} & \textbf{Framework} & \textbf{P-MNIST} & \textbf{A-ECG} & \textbf{S-EEG} & \textbf{CIFAR-100} \\ 
\midrule
SILO & SILO & 0.8774 ± 0.0088 & 0.9112 ± 0.0050 & 0.8367 ± 0.0090 & 0.4210 ± 0.0145 \\ 
\midrule
\multirow{2}{*}{CFed} 
 & BlockFL & 0.9342 ± 0.0011 & 0.9117 ± 0.0016 & 0.8358 ± 0.0045 & 0.4510 ± 0.0028 \\
 & FedMD   & 0.9375 ± 0.0017 & 0.9116 ± 0.0016 & 0.8324 ± 0.0050 & 0.4630 ± 0.0020 \\ 
\midrule
\multirow{2}{*}{SFed} 
 & pFedClub & 0.9395 ± 0.0019 & 0.9176 ± 0.0021 & 0.8421 ± 0.0077 & 0.4805 ± 0.0041 \\ 
 & SQMD   & 0.9379 ± 0.0032 & 0.9124 ± 0.0030 & 0.8446 ± 0.0063 & 0.4850 ± 0.0035 \\ 
\midrule
\multirow{2}{*}{DFed(G)} 
 & ProxyFL  & 0.9224 ± 0.0051 & 0.9051 ± 0.0066 & 0.8391 ± 0.0091 & 0.4701 ± 0.0053 \\ 
 & KD-PDFL  & 0.9232 ± 0.0033 & 0.9106 ± 0.0050 & 0.8444 ± 0.0088 & 0.4623 ± 0.0048 \\ 
\midrule
\multirow{2}{*}{DFed(W)} 
 & \bfseries TPFed-idx & 0.9375 ± 0.0023 & 0.9156 ± 0.0059 & 0.8384 ± 0.0060 & 0.4937 ± 0.0042 \\ 
 & \bfseries TPFed     & \textbf{0.9422} ± \textbf{0.0041} & \textbf{0.9287} ± \textbf{0.0038} 
                        & \textbf{0.8525} ± \textbf{0.0065} & \textbf{0.5295} ± \textbf{0.0050} \\
\bottomrule
\end{tabular}
\vspace{-15pt}
\end{table*}

\subsection{Datasets and Implementation Details}
\noindent\textbf{Datasets.} We evaluate on four benchmarks: \textbf{MNIST}~\cite{lecun1998gradient} (70{,}000 handwritten digits; the non-IID split is described under \emph{Partitioning}); \textbf{A\mbox{-}ECG}~\cite{ichimaru1999development} (35 overnight ECG recordings for sleep apnea; signals preprocessed into 60-dimensional RR-interval vectors~\cite{cai2020qrs}); \textbf{S\mbox{-}EEG}~\cite{mourtazaev1995age,rechtschaffen1968manual} (153 polysomnography recordings; we select 40 high-quality EEG records and perform three-class staging: awake, NREM, REM~\cite{rechtschaffen1968manual}); \textbf{CIFAR\mbox{-}100}~\cite{krizhevsky2009learning} (100 classes grouped into 20 superclasses; we perform within-superclass classification—e.g., flowers, insects—to increase data heterogeneity and task difficulty).

\noindent\textbf{Partitioning.} Because TPFed targets personalized analysis, multi-subject datasets adopt a “one subject = one client” scheme, yielding 35 and 40 clients for A\mbox{-}ECG and S\mbox{-}EEG, respectively. For MNIST, which lacks subject identifiers, we simulate non-IID heterogeneity by randomly dividing the training set into 20 shards and assigning two shards to each of 10 clients, while removing one digit class per shard to induce class bias. To mitigate the limited sample size in A\mbox{-}ECG and S\mbox{-}EEG, we apply a sliding-window augmentation~\cite{cai2020qrs} to each subject’s recordings. Each client’s local data are finally split into training and test sets with a $7{:}3$ ratio.

\noindent\textbf{Reference Data.} 
To enable trust-free communication without exposing private data, TPFed utilizes a shared public reference dataset $\mathbf{X}^{\text{ref}}$. In practical deployment, this data originates from publicly available repositories (e.g., open benchmarks like PhysioNet), ensuring it is non-sensitive and strictly disjoint from clients' private data. In our experiments, to simulate this setting, we reserve 20\% of the subjects from A-ECG and S-EEG to act as this public proxy data, while the remaining subjects serve as private clients. For MNIST, the standard test set (10,000 images) is used as the public reference repository. Each client uniformly samples a non-overlapping subset from this public repository to facilitate checking logits and calculating divergence.

\noindent\textbf{Training Setup.} All methods are implemented in PyTorch with a ResNet backbone~\cite{he2016identity}. The federated environment is simulated on a server with $4\times$ NVIDIA V100 GPUs. The loss $\ell$ is cross-entropy for all devices, and performance is reported as accuracy (Acc) on all datasets.

\subsection{Performance Evaluation (RQ1)} \label{sec:performance_comparison}
Table~\ref{tab:performance} reports the performance on four benchmarks. We also include \textbf{TPFed-idx}, a scalable variant detailed in Sec.~\ref{sec:scalability}, to highlight its efficiency. Overall, TPFed achieves state-of-the-art accuracy, consistently outperforming semi-decentralized and gossip-based baselines. Its superiority in heterogeneous scenarios (e.g., S-EEG, CIFAR-100) confirms the robustness of our trust-free framework. While TPFed-idx exhibits a marginal accuracy trade-off, it still surpasses existing methods.

\subsection{Resilience to LSH Cheating (RQ2)}
\begin{figure}[t] 
\centering
\vspace{0pt}
\includegraphics[scale=0.18]{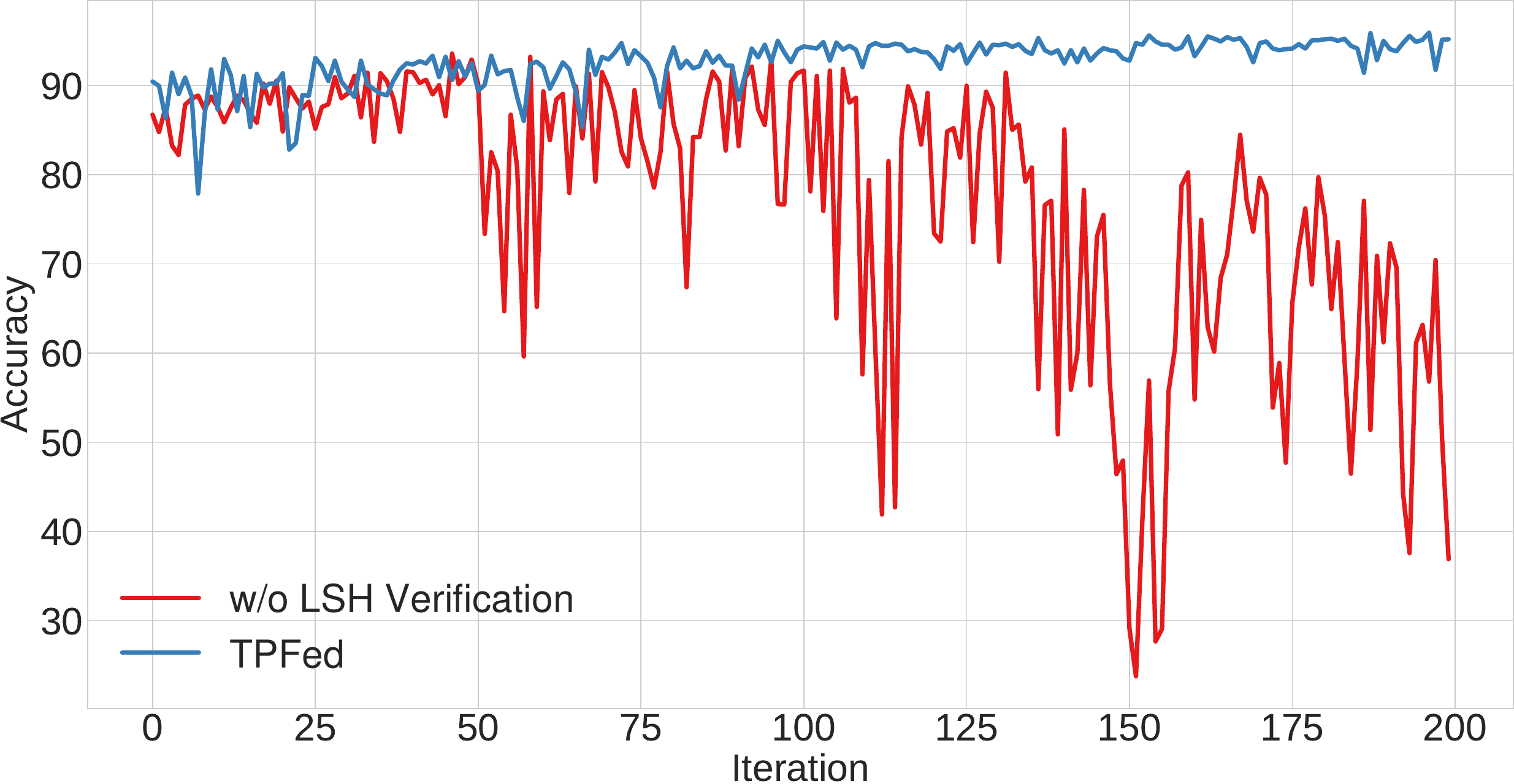}
\vspace{-5pt}
\caption{One LSH Cheating Attack Start from $t=50$.}  
\label{fig_LSH}
\vspace{-15pt}
\end{figure}

\textbf{RQ2} evaluates resilience against LSH Cheating, where adversaries forge codes to impersonate neighbors and inject malicious updates. In simulations (attack start $t=50$), Fig.~\ref{fig_LSH} shows that while unverified systems suffer sharp accuracy drops, TPFed remains stable. This robustness stems from our consistency check (Eq.~\ref{eq:consistency}): since attackers cannot generate logits semantically aligned with the victim's private reference data, their forged identities are effectively detected and rejected.

\begin{figure*}[t] 
    \centering
    
    \subfloat[]{\includegraphics[width=0.32\linewidth]{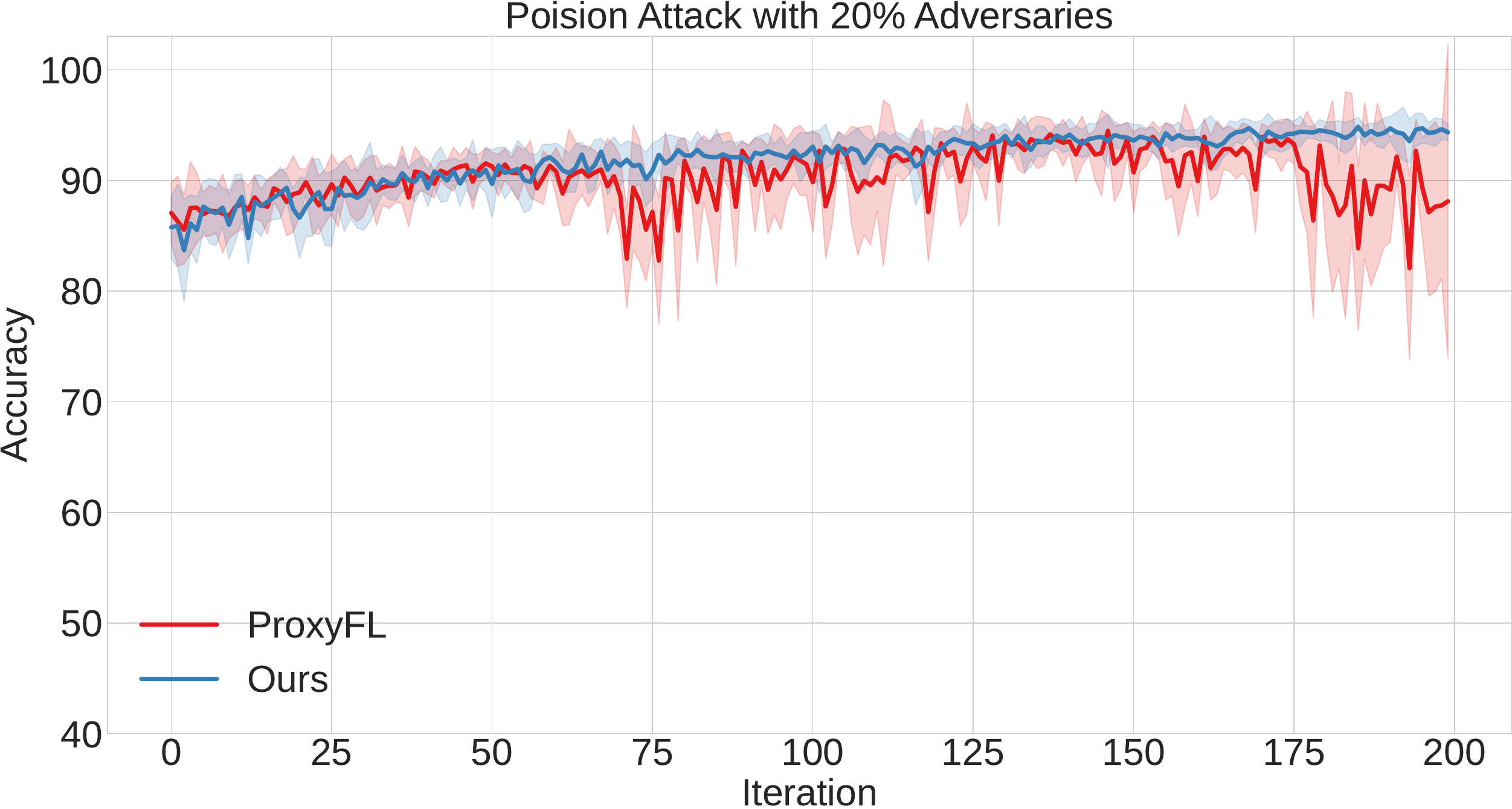}\label{fig:poison20}}
    \hfill 
    \subfloat[]{\includegraphics[width=0.32\linewidth]{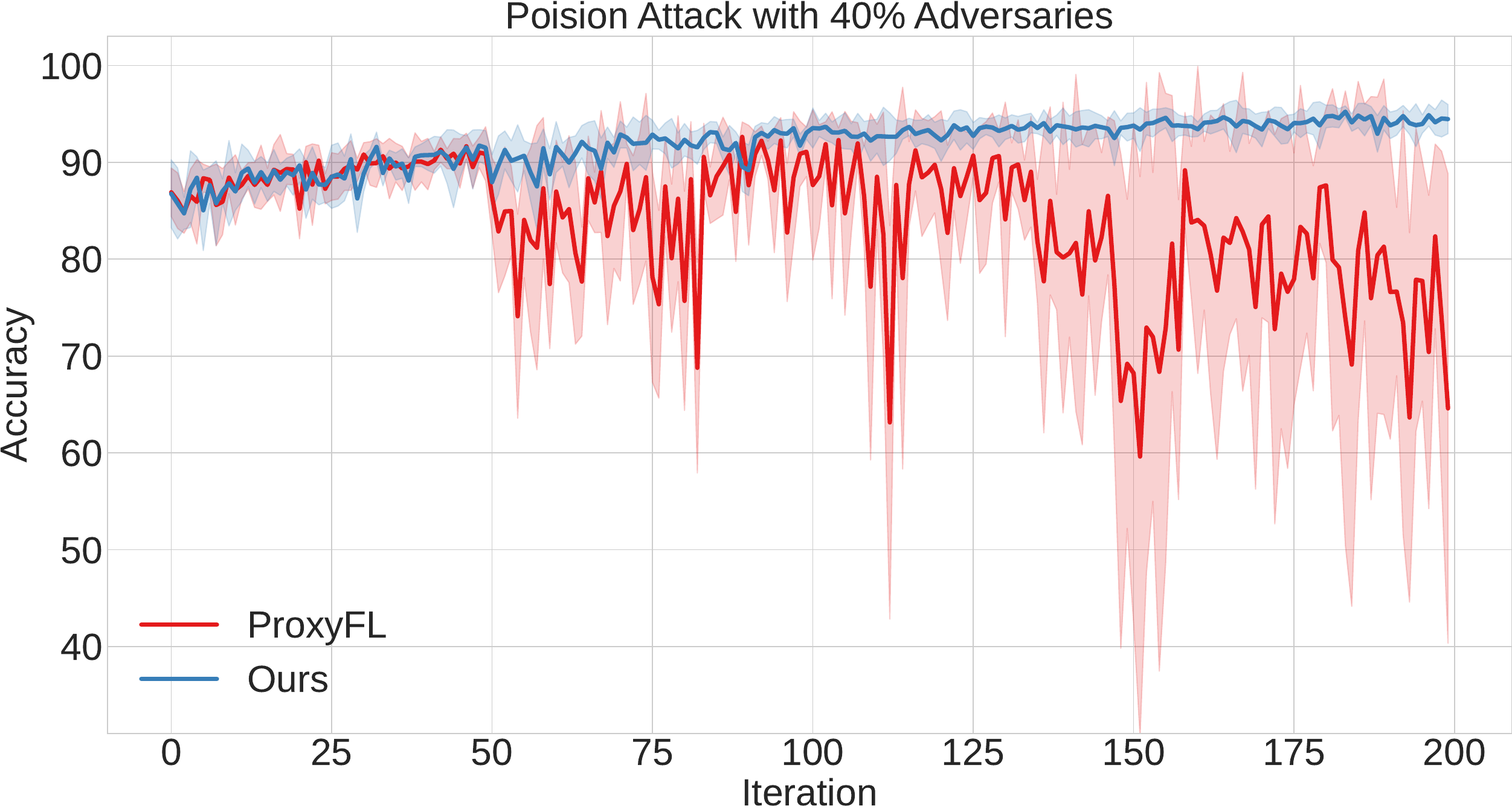}\label{fig:poison40}}
    \hfill
    \subfloat[]{\includegraphics[width=0.32\linewidth]{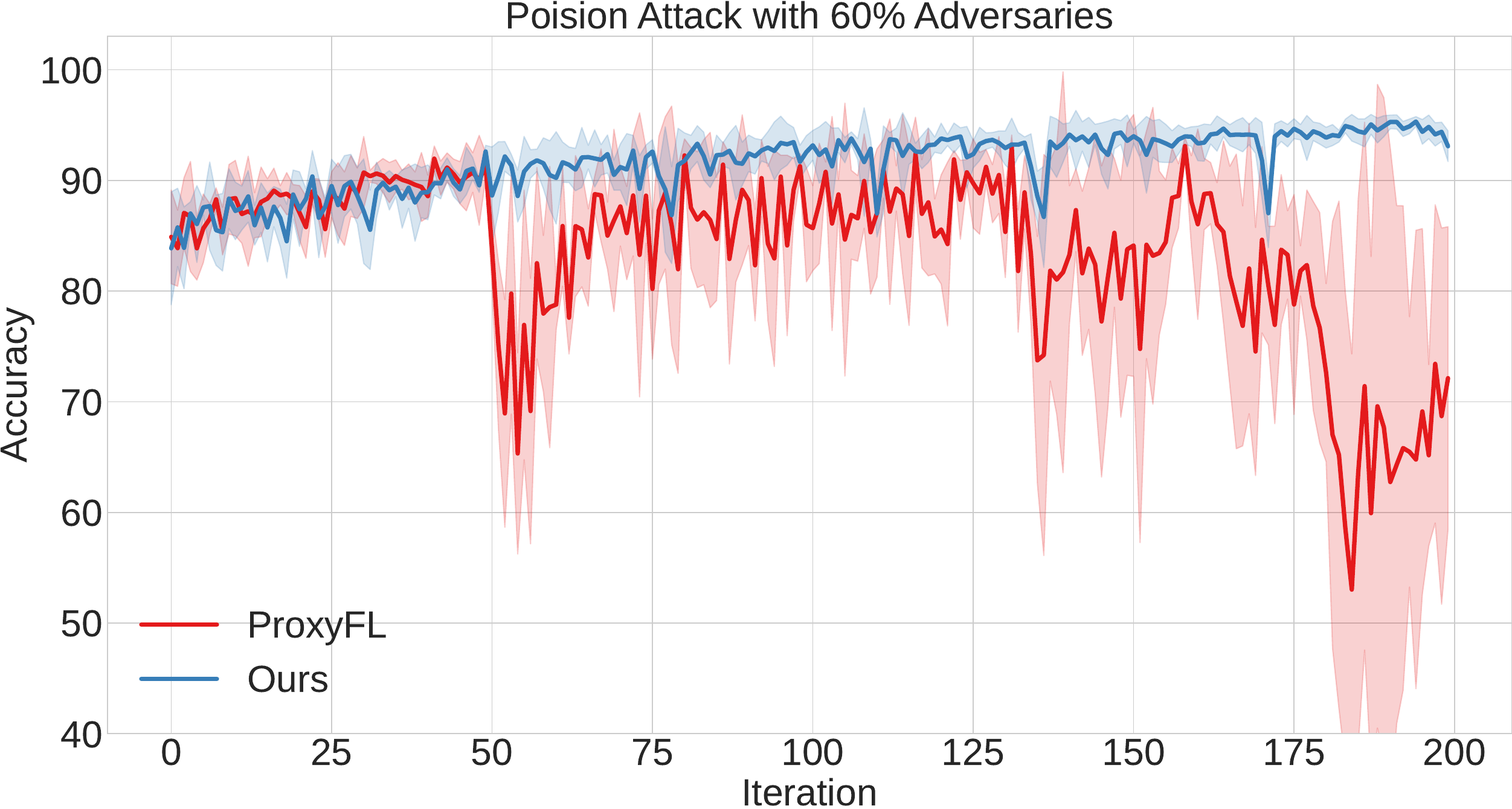}\label{fig:poison60}}

    \vspace{-2pt}
    
    \caption{Poison attacks with varying amounts of dishonest clients. Attacks start from the 50th iteration. Blue curve: TPFed; red curve: ProxyFL. Shaded areas denote the standard deviation across honest clients.}
    \label{fig:poision}
\end{figure*}

\subsection{Resilience to Poisoning Attacks (RQ3)}
We evaluated TPFed under a poisoning scenario where certain clients periodically reinitialized their model parameters after a warm-up phase, injecting noise to disrupt honest clients. Figure~\ref{fig:poision} compares the performance of TPFed and ProxyFL for varying proportions of malicious clients (20\%, 40\%, 60\%). While ProxyFL suffers notable performance drops as the fraction of malicious clients grows, TPFed remains largely unaffected at the 20\% and 40\% levels and displays only moderate fluctuations at 60\%. This resilience derives from TPFed’s global neighbor selection mechanism, wherein each client chooses high-performing (and therefore rarely malicious) peers based on both LSH similarity and rank-based evaluations, preventing malicious nodes from significantly influencing honest participants.

In all cases, the drop in accuracy remains modest, indicating that excluding delayed or missing bulletins does not drastically impair overall performance.

\subsection{Scalability (RQ4)} \label{sec:scalability}
We compare four representative methods across different categories in Table~\ref{table:overhead}. To strictly evaluate the algorithmic communication and computation costs without the volatility of public network congestion or gas fees, we conduct our simulations assuming the bulletin board is deployed on a high-throughput Layer-2 environment (as discussed in Sec.~\ref{sec_M_personal}). Under this logical infrastructure, TPFed achieves a convergence speed on par with gossip-based decentralized learning but exhibits a higher Extra Overhead per Round (1.25$\times$) compared to baselines.

\begin{table}[t]
\vspace{-0pt}
\centering
\caption{Overhead of four representative methods (Set the overhead of FedMD as the baseline).}
\label{table:overhead}

\footnotesize
\renewcommand{\arraystretch}{1.1}

\begin{tabularx}{\linewidth}{YYYY}
\toprule
\textbf{Category} & \textbf{Method} &
\makecell{\textbf{Overhead}\\\textbf{per Round}} &
\makecell{\textbf{Overhead}\\\textbf{at 40\% Acc}} \\
\midrule
CFed   & FedMD   & 1.00 & 1.00 \\
SFed   & SQMD    & 1.05 & 1.10 \\
DFed(G)& ProxyFL & 1.08 & 1.22 \\
DFed(W)& TPFed   & 1.25 & 1.21 \\
\bottomrule
\end{tabularx}
\vspace{-10pt}
\end{table}

Therefore, we present a detailed theoretical analysis and introduce a scalable variant, \textbf{TPFed-idx}. Unlike the standard TPFed where every client parses the entire bulletin board, TPFed-idx keeps the training process decentralized while leveraging the blockchain’s global index as an \emph{auxiliary service}. A smart contract pre-sorts the LSH codes to generate a global index, allowing clients to retrieve only the top-$H$ relevant candidates ($H \ll M$). This reduces the client-side sorting complexity significantly. We evaluate both TPFed and TPFed-idx on CIFAR-100 and summarize the results in Table~\ref{table:compare}.

\begin{table}[t]
\centering
\caption{Comparison of TPFed and TPFed-idx with H=20 on CIFAR-100. (M is the number of clients. H is the retrieve range. Overhead per round is normalized to FedMD)}
\label{table:compare}

\footnotesize
\renewcommand{\arraystretch}{1.1}

\begin{tabularx}{\linewidth}{YYYYY}
\toprule
\textbf{M} &
\makecell{\textbf{TPFed}\\\textbf{Acc}} &
\makecell{\textbf{TPFed}\\\textbf{Overhead}} &
\makecell{\textbf{TPFed-idx}\\\textbf{Acc}} &
\makecell{\textbf{TPFed-idx}\\\textbf{Overhead}} \\
\midrule
20  & 0.4505 & 1.17 & 0.4505 & 1.17 \\
100 & 0.5295 & 1.25 & 0.4937 & 1.17 \\
500 & 0.5323 & 1.68 & 0.4991 & 1.17 \\
\bottomrule
\end{tabularx}
\vspace{-15pt}
\end{table}

\subsection{Ablation Study (RQ5)}
We conduct an ablation study to evaluate the contributions of each main component in \textbf{TPFed} by removing them individually. Specifically, we consider three variants: \textbf{w/o LSH} discards the locality-sensitive hashing approach, relying entirely on peer ranking (PR) to select neighbors. \textbf{w/o PR} excludes the ranking mechanism, basing neighbor selection solely on LSH-derived similarity. \textbf{w/o logits} removes the all-in-one logits exchange, so neither LSH nor PR can be verified during P2P interactions.

Table~\ref{tab_ablation} reports the performance differences of these variants relative to full TPFed. Omitting either LSH or PR leads to noticeable accuracy drops (about 1–3\% across most datasets), confirming that both similarity checks and peer ranking are crucial for effective neighbor selection. Notably, \textbf{w/o logits} suffers the largest decrease in performance, highlighting the importance of verifying claims about model quality and similarity through the reference-data logits. This verification mechanism directly prevents adversarial behaviors and ensures that LSH codes and PR scores accurately reflect genuine model characteristics.

\begin{table}[t]
\vspace{-15pt}
\centering
\caption{Ablation Study for TPFed}
\label{tab_ablation}

\footnotesize
\renewcommand{\arraystretch}{1.1}

\begin{tabularx}{\linewidth}{YYYYY}
\toprule
\textbf{Config} & \textbf{MNIST} & \textbf{A-ECG} & \textbf{S-EEG} & \textbf{CIFAR} \\
\midrule
\textbf{TPFed}   & \textbf{0.9422} & \textbf{0.9287} & \textbf{0.8525} & \textbf{0.5295} \\
w/o LSH          & -0.0104         & -0.0070         & -0.0153         & -0.0194 \\
w/o PR           & -0.0116         & -0.0061         & -0.0198         & -0.0308 \\
w/o logits       & -0.0282         & -0.0195         & -0.0202         & -0.0371 \\
\bottomrule
\end{tabularx}
\vspace{-5pt}
\end{table}

\subsection{Abnormal Uploads (RQ6)}
Although TPFed uses blockchain-based bulletins for decentralized coordination, real-world deployments may encounter latency, network congestion, or node misbehavior. We incorporate a \emph{timeout} mechanism: if a client’s bulletin fails to appear on-chain before a certain deadline, other clients skip its data in that round. Table~\ref{tab:delay-minimal} shows three scenarios where $20\%$ or $40\%$ of clients delay their uploads beyond 50\% of the median latency, and a case where $20\%$ of clients fail to upload entirely.

\vspace{-5pt}
\begin{table}[t]
\centering
\caption{Impact of delayed and failed uploads on TPFed’s accuracy.}
\label{tab:delay-minimal}
\small
\renewcommand{\arraystretch}{0.9}
\begin{tabularx}{\linewidth}{YYYY}
\toprule
\textbf{Dataset} &
\textbf{20\% Delay} &
\textbf{40\% Delay} &
\textbf{20\% Fail} \\
\midrule
MNIST & -0.0085 & -0.0132 & -0.0143 \\
A-ECG & -0.0063 & -0.0091 & -0.0120 \\
S-EEG & -0.0049 & -0.0078 & -0.0101 \\
CIFAR & -0.0024 & -0.0042 & -0.0065 \\
\bottomrule
\end{tabularx}
\vspace{-15pt}
\end{table}

\subsection{Low-Quality Reference Data (RQ6)}

Although TPFed relies on each client’s reference dataset to evaluate and select neighbors, real-world data often contain biases, noise, or shifts in distribution. To assess TPFed’s resilience, we conducted three experiments where varying proportions (20\%, 40\%, and 60\%) of clients replaced part of their reference data with distorted or adversarial samples. For instance, in one setup on CIFAR, local noise (potentially privacy-threatening) was injected to simulate suboptimal reference images. Table~\ref{tab:biased-ref} reports the resulting accuracy drops ($\Delta \text{Accuracy}$) compared to a baseline with clean, unbiased reference data.

\begin{table}[t]
\vspace{-25pt}
\centering
\caption{Performance impact from biased or low-quality reference data across different corruption rates.}
\label{tab:biased-ref}

\footnotesize
\renewcommand{\arraystretch}{1.1}

\begin{tabularx}{\linewidth}{YYYY}
\toprule
\textbf{Dataset} &
\textbf{20\% Corr.} &
\textbf{40\% Corr.} &
\textbf{60\% Corr.} \\
\midrule
MNIST & -0.0066 & -0.0104 & -0.0189 \\
A-ECG & -0.0011 & -0.0023 & -0.0055 \\
S-EEG & -0.0005 & -0.0013 & -0.0037 \\
CIFAR & -0.0008 & -0.0015 & -0.0024 \\
\bottomrule
\end{tabularx}
\vspace{-15pt}
\end{table}

Overall, TPFed maintains robust performance even under 40\% corruption, with an accuracy reduction of less than 2\% for most datasets. At 60\% corruption, some additional degradation becomes apparent, yet the final accuracy remains competitive. This resilience arises from TPFed’s multi-level checks—clients not only evaluate peers with reference-data logits (offering a direct performance metric) but also use LSH and rank-based verification to filter out untrustworthy neighbors. However, in extreme cases where reference data are thoroughly compromised or absent, TPFed’s all-in-one approach cannot function properly. To mitigate such scenarios, reference samples can be derived from more diverse data sources, augmented periodically, or verified cooperatively across multiple clients. Empirically, our experiments confirm TPFed’s tolerance of moderate biases, ensuring stable personalization and training in real-world deployments. 


\section{Conclusion}
This study proposes TPFed, a fully decentralized federated learning framework that integrates inter-client similarity and model quality assessments for neighbor selection, coupled with privacy-preserving knowledge distillation and LSH verification mechanisms. TPFed enhances learning performance, system robustness, and data privacy, making it a promising solution for secure and personalized federated learning in sensitive applications such as e-health and decentralized finance.

\textbf{Limitations:} First, regarding model compatibility, TPFed currently operates best under homogeneous architectures with shared initialization. While neural network permutation invariance may theoretically affect LSH accuracy, our experiments suggest that even in cases where permutation leads to falsely large distances, TPFed's redundancy allows clients to find other sufficient neighbors, while the subsequent logits verification prevents false positives. Second, regarding system incentives, the blockchain component currently lacks granular punitive mechanisms to penalize malicious actors beyond simple exclusion. Future work will focus on extending TPFed to heterogeneous model settings and integrating game-theoretical incentive designs.




\section*{Acknowledgments}
This work was supported by the National Natural Science Foundation of China (U22B2038, 62192784), the Fundamental Research Funds for the Central Universities (510224045), the National Natural Science Foundation of China (U23A20319, 62272054), and the Beijing Nova Program (20230484319).

\bibliographystyle{IEEEtran}
\bibliography{example_paper}



\vfill

\end{document}